\documentclass[journal]{IEEEtran}

\usepackage{times}
\usepackage{epsfig}
\usepackage{graphicx}
\usepackage{amsthm}
\usepackage{amsmath}
\usepackage{amssymb}
\usepackage{subfigure}
\usepackage{epstopdf}
\usepackage{algorithm}
\usepackage{algorithmic}
\usepackage{multirow}
\usepackage{color}
\usepackage{ragged2e}
\usepackage{caption}
\usepackage[hyphens]{url}
\PassOptionsToPackage{hyphens}{url}
\usepackage{breakurl}
\usepackage{footnote}

\DeclareMathOperator*{\argmin}{argmin}

\theoremstyle{plain}

\ifCLASSINFOpdf
\else
\fi
\begin{document}
%
\title{Automatic Image Cropping for Visual Aesthetic Enhancement Using Deep Neural Networks and Cascaded Regression}
%
%
%

\author{Guanjun~Guo,
        Hanzi~Wang*,~\IEEEmembership{Senior Member,~IEEE,}
        Chunhua~Shen,~\IEEEmembership{}
        Yan~Yan,~\IEEEmembership{Member,~IEEE,} \\
        Hong-Yuan~Mark Liao,~\IEEEmembership{Fellow,~IEEE}
\IEEEcompsocitemizethanks{
\IEEEcompsocthanksitem *~Corresponding author. Tel./fax: +86 5922580063.\protect
\IEEEcompsocthanksitem G.~Guo, H.~Wang and Y.~Yan are with the Fujian Key Laboratory of Sensing and Computing for Smart City, and the School of Information Science and Engineering, Xiamen University, Xiamen 361005, Fujian, P. R. China.\protect
~(E-mail: gjguo@stu.xmu.edu.cn; hanzi.wang@xmu.edu.cn; yanyan@xmu.edu.cn).
\IEEEcompsocthanksitem C. Shen is with the Australian Center for Visual Technologies, and the School
of Computer Science at The University of Adelaide, SA 5005, Australia.\protect
 ~(E-mail: chunhua.shen@adelaide.edu.au).
\IEEEcompsocthanksitem H. Y. Mark Liao is with the Institute of Information
Science, Academia Sinica, Taipei 115, Taiwan.\protect
 ~(E-mail: liao@iis.sinica.edu.tw).
 }
\thanks{}}

%
%

\markboth{IEEE TRANSACTIONS ON MULTIMEDIA}%
{Shell \MakeLowercase{\textit{et al.}}: Bare Demo of IEEEtran.cls for IEEE Journals}
%



\maketitle

\begin{abstract}
Despite recent progress, computational visual aesthetic is still challenging. Image cropping, which refers to the removal of unwanted scene areas, is an important step to improve the  aesthetic quality of an image.  However, it is challenging to evaluate whether cropping leads to aesthetically pleasing results because the assessment is typically subjective. In this paper, we propose a novel cascaded cropping regression (CCR) method to perform image cropping by learning the knowledge from professional photographers. The proposed CCR method improves the convergence speed of the cascaded method, which directly uses random-ferns regressors. In addition, a two-step learning strategy is proposed and used in the CCR method to address the problem of lacking labelled cropping data. Specifically, a deep convolutional neural network (CNN) classifier is first trained on large-scale visual aesthetic datasets. The deep CNN model is then designed to extract features from several image cropping datasets, upon which the cropping bounding boxes are predicted by the proposed CCR method. Experimental results on public image cropping datasets demonstrate that the proposed method significantly outperforms several state-of-the-art image cropping methods.
\end{abstract}


\IEEEpeerreviewmaketitle

\section{Introduction}
\label{sec:introduction}
\IEEEPARstart{C}{omputational} image understanding consists of not only image classification, object detection, object tracking and other popular computer vision tasks, but also semantic inference of aesthetic from images. Computational inference of aesthetic has a wide range of applications. Image aesthetics can be used to recommend  aesthetically pleasing images in an image repository to users. General consumers or designers can utilize the feedback from an automated aesthetic evaluation system to improve decisions \cite{Aesthetic_Emotion2011}. However, computational inference of image aesthetics  remains a challenging task because image aesthetics is highly subjective and difficult to represent with precise mathematical explanations.

\begin{figure}[t]
\begin{center}
   \includegraphics[width=0.98\linewidth]{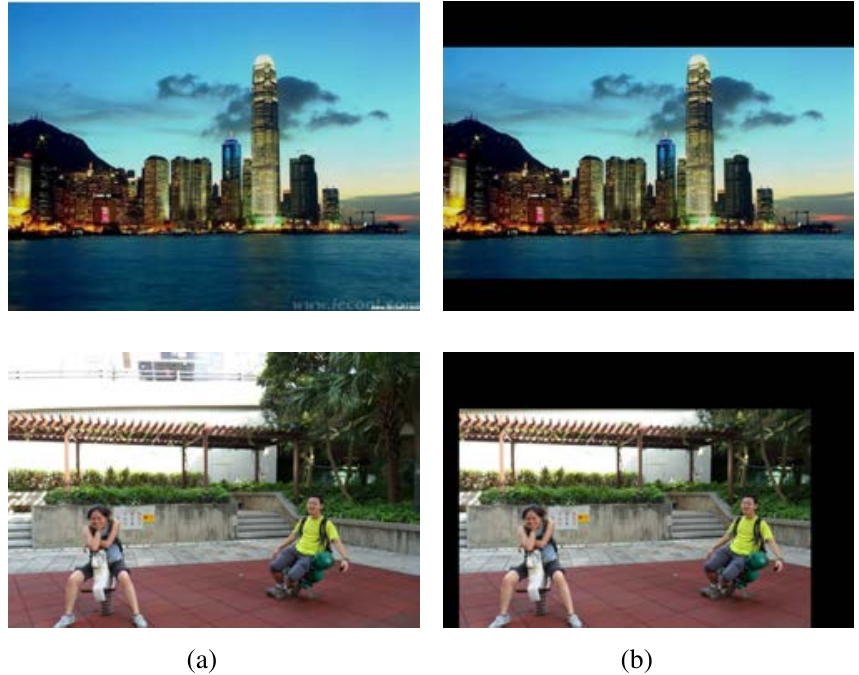}
\end{center}
\vspace{-0.391cm}
   \caption{Illustration of the proposed CCR method to crop aesthetic regions in two images. (a) Input images. (b) The cropped images obtained by the proposed method. The black regions are cut off by the proposed method.}
\label{Fig::demo}
\end{figure}
Although aesthetic measurement is subjective, some works attempt to learn visual properties that make photographs aesthetically beautiful~\cite{StudyAesthetics2006,MarchesottiPLC2011,Yan2013,zhangtmm2014,MM14,AliVA15}. For example, in~\cite{StudyAesthetics2006}, color, texture and other low-level visual features are extracted for training a two-class classification tree model to predict the aesthetic quality of an image. Several recent methods~\cite{MarchesottiPLC2011,Yan2013,tang2013} have followed this scheme. Specifically, different handcrafted features inspired by aesthetic experience are first extracted. Machine learning models are then trained to identify the visual properties of most relevance to the aesthetic quality. As deep CNN~\cite{alexNet}  is successful on many computer vision tasks, where features are automatically learned from raw images, CNN has also been used on aesthetic categorization for images~\cite{RAPID2014,mai2016,deepSim}. Typically, CNN requires a large number of training samples to alleviate overfitting. A recently published large-scale dataset called the large-scale aesthetic visual analysis (AVA) dataset~\cite{AVA2012} makes it feasible to train a CNN based aesthetic classifier. The task of blind image quality assessment (BIQA) is closely related to the task of aesthetic categorization, and BIQA methods (such as~\cite{gaofei2017,biqa2011}) do not need many training samples, these BIQA methods usually focus high-quality image regions rather than object regions. Thus, these BIQA methods are usually not appropriate for the task of image cropping.

Image cropping is a key step to generate aesthetically pleasing images. A distinction between a professional photographer and an amateur heavily relies on whether one can remove distracting contents and highlight desired subjects to enhance the visual aesthetic of an image. Some composition rules for obtaining aesthetically pleasing photographs, such as the rules of thirds (i.e., an approximation of the Golden Ratio) and the rules of odds have been developed. Inspired by these composition rules, several image cropping methods, which are usually based on handcrafted features, have been proposed. Representative works are \cite{Cheng2010,Yan2013}. However, the composition rules may be different for various styles of images. For example, portrait photographers often highlight a person while landscape photographers focus more on the interaction among the elements of an image. In Gestalt psychology~\cite{koffka1935}, the concept of goodness of configuration shows that the elements in an image are not isolated. People prefer to choose patterns, which have properties such as symmetry, simplicity, etc. Therefore, image cropping for visual aesthetic enhancement is a complicated and high-level cognition process.

In this paper, we develop a machine learning method to automatically crop images to produce the cropped images that are aesthetically pleasing. Specifically, a two-step learning approach is proposed to solve the automatic cropping problem with a small number of training samples that contain cropping information. First, a CNN classifier is trained using a large collection of visual aesthetic datasets. The CNN features are then extracted based on the trained classifier. Using the CNN features as the input, a cascaded cropping regression (CCR) method, which combines a set of weak  random-ferns regressors \cite{cascadedRe2010} as a primitive regressor, is proposed to fit the image cropping information annotated by professional photographers.
A key motivation of this two-step learning approach is that the amount of image cropping data is limited and directly fitting a cropping model using deep CNNs may easily lead to overfitting. Bounding-box labels are very limited and expensive, especially in the case where bounding-box labels are annotated by professionals. However, a large number of weakly labeled images, which have image-level aesthetic labels, are available from the internet. Therefore, this work presents a two-step learning approach to respectively learn a CNN feature extractor and a set of regressors for the task of image cropping. The advantages of the proposed two-step learning approach are as follows. First, the proposed approach leverages a large number of aesthetic image data, and thus can effectively learn image features. Second, the proposed CCR method is effective in selecting features and learning cropping information in a cascaded manner. At each stage of regression, CNN features are extracted from the cropping region obtained at the previous stage. The final cropping result is obtained after several stages of regression. The primitive regressor employed in the CCR method is an ensemble of random-ferns regressors selected by using the gradient boosting algorithm~\cite{Friedman01}. Each primitive regressor indirectly selects features from the CNN features by aggregating the features selected by the set of random-ferns regressors. In contrast to conventional cascaded methods, which directly use one random-ferns regressor as the primitive regressor, the CCR method converges quickly in a few stages, significantly reducing the computational complexity.

To the best of our knowledge, this work is the first to present a cascaded regression method with the CNN features for the task of automatic image cropping.  We improve the convergence speed of the cascaded regression method based on random-ferns regressors, which makes the proposed image cropping method quite efficient. In addition, we propose a two-step learning strategy for limited labelled cropping data. The proposed image cropping method is  effective and significantly outperforms several state-of-the-art image cropping methods. Fig.~\ref{Fig::demo} illustrates the proposed image cropping method for visual aesthetic enhancement. The proposed method can imitate the manipulations of expert photographers to remove distracted regions (such as watermarks, non-subject regions, etc.) and highlight a subject in an image.

\section{Related Work}
\label{sec::relatedW}
Existing image cropping methods can be roughly divided into three categories. Methods of the first category are attention-based, and output a cropping bounding box around an informative object. The informative object can be a salient object obtained by different saliency detection methods (such as~\cite{saliencyFilters2012,guo2017}). For example, Marchesotti et~al.~\cite{Marchesotti2009} propose a framework for visual saliency detection where one or more thumbnails are extracted from the obtained saliency maps. Thumbnails are usually salient foreground regions while non-informative pixels become part of the background. Fang~et~al.~\cite{Fang2014} also utilize a spatial pyramid of saliency maps as a composition feature to force a cropped image to contain a salient object. In addition, faces or other regions of interest are often used as informative regions for image cropping~\cite{Ciocca,VikramTWPR12,Laurentini2014184}.

\begin{figure*}[t]
\centering
   \includegraphics[width=0.98\linewidth]{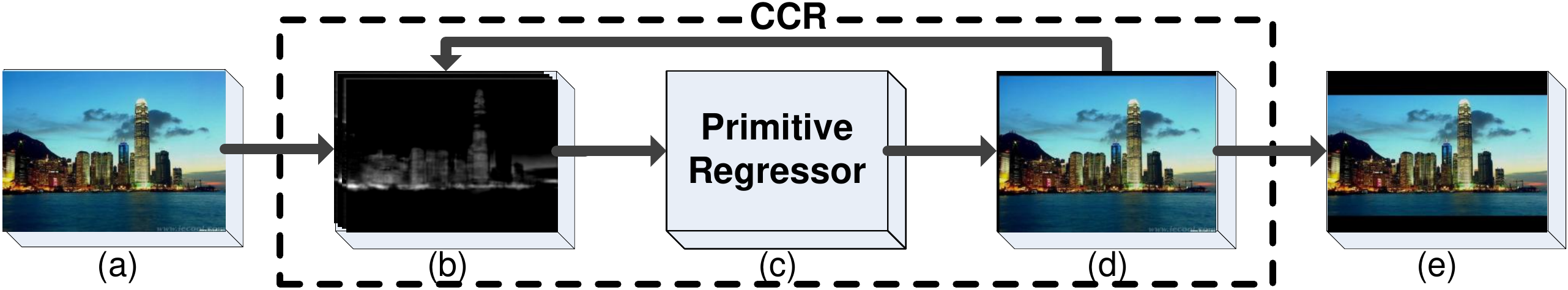}
   \caption{The framework of the proposed CCR method. The dashed box denotes the procedure where $T$ stages of regression are performed for obtaining the final cropping result. (a) An input image (i.e., the initial cropping region). (b) The cropping-indexed CNN features. (c) The proposed primitive regressor used to predict the values of the cropping region increment based on the features obtained in (b). (d) Updating the cropping region. (e) The final cropping result.}
\label{Fig:framework}
\end{figure*}
Methods of the second category rely on the aesthetic evaluation of cropping results. Machine learning methods are often used to identify the aesthetic of cropped images. Moreover, these methods also consider the spatial distribution of elements in an image to obtain an optimized arrangement of the image elements. Representative works are \cite{Nishiyama2009,Cheng2010,Zhang2013}. However, aesthetic based methods are sensitive to evaluating the attractiveness of a cropped image, and therefore these methods focus on what remains in the cropped image. To overcome the above problem, Yan~et~al.~\cite{Yan2013,Yan2015} propose an image cropping method belonging to the third category of image cropping methods, namely an experience-based method. In their method, they construct several cropping datasets, which are annotated by three professional photographers, for image cropping. Various  handcrafted features are then extracted for regressing the cropping values annotated by the professional photographers. This method emphasizes professionals' experience or the change caused by the manipulations of image cropping.  A drawback is that handcrafted features are often limited  and may lack useful information for high-level computer vision tasks.

Inspired by the success of deep CNNs on various vision tasks,  in this paper we propose to learn the CNN features for automatic image cropping. To alleviate overfitting, the CNN model is trained on a combination of the large-scale AVA dataset and the CHUKPQ dataset. The learned model is then applied to several cropping datasets for extracting the CNN features. With the extracted CNN features, we propose the CCR (cascaded cropping regression) method to fit the cropping information annotated by professional photographers.

\section{The Proposed CCR Method}
This section provides the details of the proposed CCR method for visual aesthetic enhancement. Fig.~\ref{Fig:framework} shows the framework of the proposed CCR method. The size of the initial cropping region $C^0$ is set to be the same as that of the original image $I$. The CNN features are extracted from the cropping region $C^0$ by using a pre-trained CNN model (the training details of the CNN model will be presented in subsection~\ref{sec::training}). Then a primitive regressor is used to estimate an improved cropping region using the  CNN features at each stage. The final cropping result is obtained after $T$ stages of regression.

A cropping region $C=[\hat{x}_1,\hat{y}_1,\hat{x}_2,\hat{y}_2]^T$ consists of the coordinates of the top-left and bottom-right corners of a rectangle region. Given an original image $I$, the goal of image cropping is to estimate a cropping region $C$ that is as close as possible to the ground-truth cropping region $\hat{C}$ provided by a professional photographer. The above problem can be modeled as the least-squares regression problem  as follows:
 \begin{equation}
  \bar{C}=\mathop{\argmin}_{C}{\|C-\hat{C}\|^2}.
\label{Eq::regCost1}
\end{equation}
This least-squares regression problem\footnote{The least-squares loss function shown in Eq.~(\ref{Eq::regCost1}) can be improved by using the other loss functions (such as~\cite{wang2014} and ~\cite{yu2017}).} is solved by the proposed CCR method. Before describing the proposed CCR method in detail, we propose a CNN feature extraction strategy in the next subsection.

\subsection{Training an Aesthetic Classifier with CNN for Feature Extraction}
\label{sec::training}
Although some handcrafted features (such as exclusion and compositional features in \cite{Yan2013,Yan2015}) for image cropping have been proposed, these features fail to cover all possible situations for image cropping. Moreover, visual aesthetic assessment is subjective. Therefore, instead of designing some handcrafted features, the proposed method directly learns features
from large-scale visual aesthetic datasets based on the deep convolutional neural network (CNN). The AVA dataset~\cite{AVA2012} and the CUHKPQ dataset~\cite{Luo2011} are two annotated large-scale visual aesthetic datasets, where most images are from the photography website of Dpchallenge\footnote{http://www.dpchallenge.com}. The images of the two datasets received a number of votes or aesthetic judgements from the members of the photography community. As high-quality visual pleasing images usually have a clear topic and an impressive composition, the features learned from these images are relatively more useful for enhancing visual aesthetic during the image cropping process.  Note that a cropping regressor cannot be directly trained using CNN on these two datasets due to the lack of cropping annotation.

\begin{figure*}[t]
  \centering
   \includegraphics[width=0.95\linewidth]{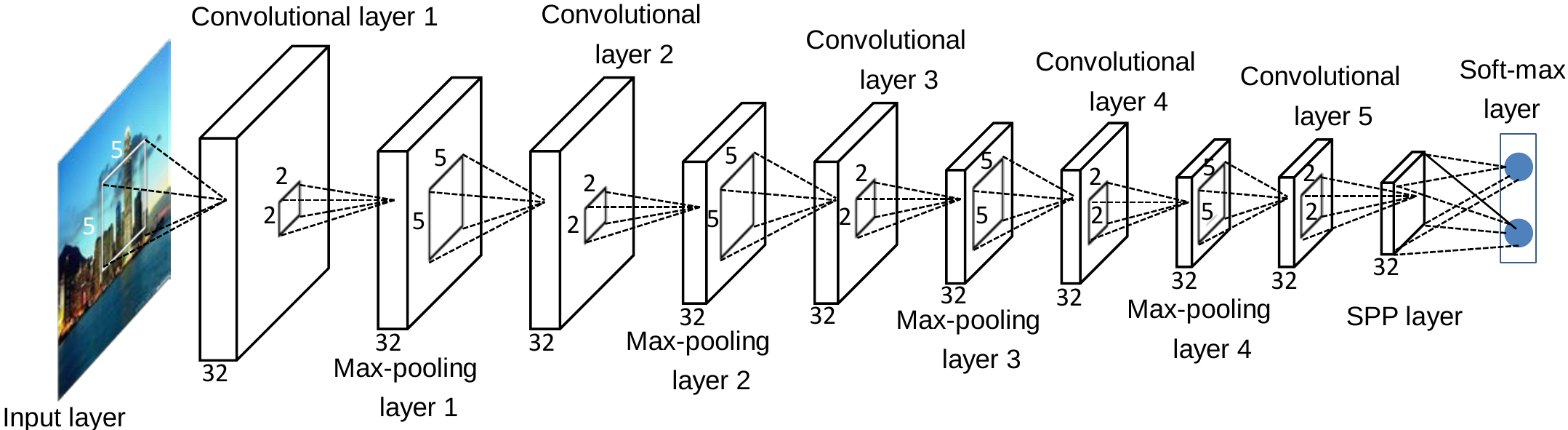}
 \caption{The proposed deep CNN structure, which is trained using the combined AVA and CHUKPQ datasets, is used to extract the CNN features.}
 \label{Fig:CNNStructure}
 \end{figure*}

We design a deep CNN structure (as shown in Fig.~\ref{Fig:CNNStructure}) to extract valid features for image cropping. The first convolutional layer is used to filter input images with $32$ kernels of size $5\times5\times3$ and it outputs 32 feature maps, which have the same size as each of the input images.  Each convolutional layer also outputs 32 feature maps.
Each convolutional layer is then activated by a rectified linear unit (ReLU)~\cite{ReLU2013} and followed by a max-pooling layer in the first four layers. The max-pooling layer partitions the feature maps from the previous layer into a set of $2\times2$ non-overlapping neighborhoods, and outputs the maximum value for each neighborhood. The last layer is the spatial pyramid pooling layer (SPP)~\cite{sppnet2014}, which partitions each input feature map into divisions from a fine level to a coarse level and aggregates the local features in each division. Using the SPP layer can alleviate the problem that  resizing a visually pleasing image may potentially damage its aesthetic. In total, five convolutional layers, four max-pooling layers and one SPP layer are used in the designed network structure (as shown in Fig.~\ref{Fig:CNNStructure}). The CNN model corresponding to this structure is trained on the combined AVA and CUHKPQ datasets.

\textbf{Training datasets.}
The aesthetic classifier is trained using the combined AVA and CUHKPQ dataset. The AVA dataset contains more than 250,000 images collected from the Dpchallenge website. Each image has about 210 aesthetic ratings ranging from $1$ to $10$. We divide these images into two categories (i.e., low-quality images and high-quality images) for training a two-class CNN model. Following the same strategy used in \cite{AVA2012,RAPID2014}, a parameter $\delta$ is used to discard ambiguous images from the training set. The images with average scores smaller than $5-\delta$ are referred to as low-quality images. The images with average scores larger than or equal to $5+\delta$ are considered as high-quality images. The images with average scores between $5-\delta$ and $5+\delta$ are considered as ambiguous and are thus discarded. In the implementation, we set the value of $\delta$ to be 1. Thus, 49,682 high-quality images and 7,983 low-quality images are selected from the AVA dataset. The CHUKPQ dataset contains about 30,000 images, which are collected from a variety of photography websites. Each image in this dataset has been labelled as either low or high quality. There are totally 10,525 high-quality images and 19,167 low-quality images labelled in the CHUKPQ dataset. The images from the two datasets are combined according to their categories. In total, 60,207 high-quality images and 27,150 low-quality images are collected from these two datasets. To alleviate the class imbalance problem,  low-quality images are augmented  by flipping each low-quality image horizontally to obtain 54,300 low-quality images in total. All the obtained images are split into 894 batches, and each batch consists of 128 images.

\begin{figure*}[t]
\centering
    \centerline{\includegraphics[width=0.98\textwidth]{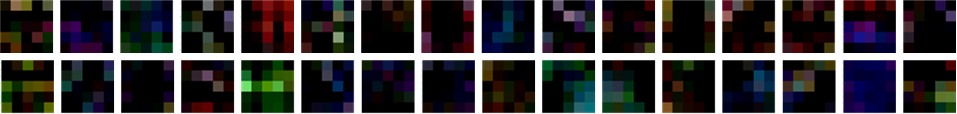}}
   \caption{The trained filters of the first convolutional layer of the proposed CNN model for image aesthetic classification.}
\label{Fig:CNNFilter}
\end{figure*}
\begin{figure*}[t]
\centering
   \centerline{\includegraphics[width=0.98\textwidth]{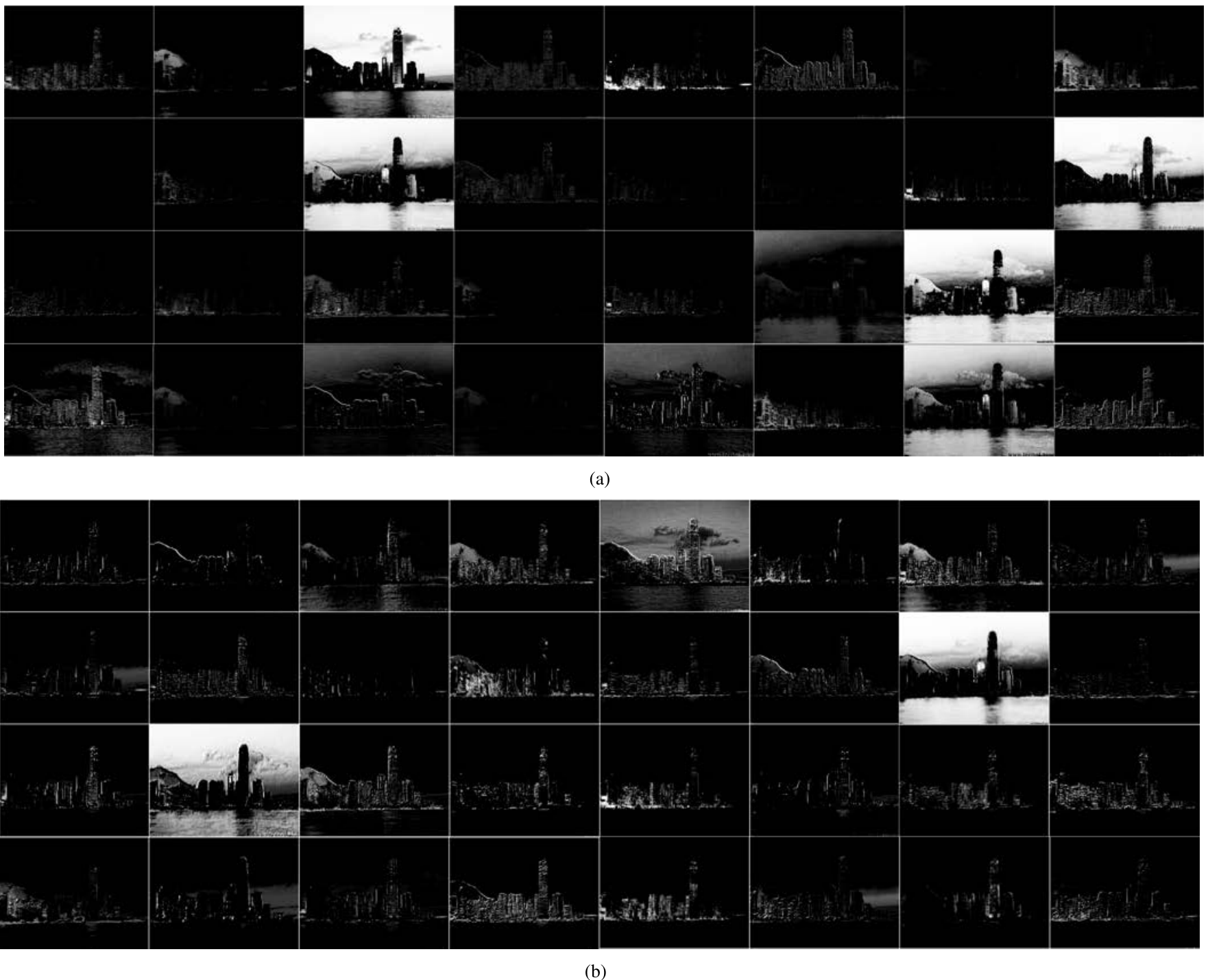}}
   \caption{The feature maps obtained by the proposed CNN model on the image shown in the first row in Fig.~\ref{Fig::demo} (a). (a)-(b) The feature maps obtained in the first and second convolutional layers of the CNN model, respectively.}
\label{Fig:CNNFeat}
\end{figure*}
\textbf{Training the aesthetic classifier.}
In this paper, we use the GPU implementation code of $Torch7$~\cite{Collobert_torch7} to train the CNN model. 880 batches are used for training and 14 batches are used for testing. The whole training process takes around one day on a computer equipped with two GTX-1080 GPUs, and the trained CNN achieves 75.1\% classification accuracy in test. This result is competitive even compared with that obtained by the state-of-the-art aesthetic classification method~\cite{RAPID2014}. A part of the filters in the trained model are shown in Fig.~\ref{Fig:CNNFilter}. Most filters in the first convolutional layer are related to color. Fig.~\ref{Fig:CNNFeat} shows the feature maps obtained by the CNN model on the image shown in the first row of Fig.~\ref{Fig::demo}.

The trained CNN model is then applied on each image in the cropping datasets. The feature maps before the classifier layer (i.e., the feature maps in the SPP layer in Fig.~\ref{Fig:CNNStructure}) are obtained and used for cropping regression. In our implementation, the SPP layer has three levels, where the feature maps of these three levels are respectively partitioned into $2\times2$, $3\times3$ and $4\times4$ divisions. An input image becomes feature maps with the size of $(2\times2+3\times3+4\times4)\times32$ after $5$ layers of pooling. Thus, the dimension of the obtained features is $928$. Let $t$ denote the index of the stage in the proposed CCR method. In the $t$th stage of regression, the CNN features $x_i^t$ are extracted from the cropped image obtained from the ($t-1$)th  stage of regression. We term these CNN features as the cropping-indexed CNN features. Such features correlate with spatial positions and also significantly improve the cascaded regression. However, because the number of training samples for anesthetic image cropping is often limited, existing regressors (such as CNN and random-fern regressors) cannot be  used directly in this task. In the following subsection, we propose a novel and effective regressor to address the problem caused by a limited number of training data for the task of image cropping.

\subsection{Learning a Primitive Regressor using Gradient Boosting Algorithm}
\label{sec::learingPR}

Inspired by the cascaded regression methods used in the tasks of pose regression~\cite{cascadedRe2010} and face alignment~\cite{cao2014}, a cascaded regressor $R_j=(r_j^1,\cdots,r_j^T)$ is trained to output $C_j$, where $j$ ($1\leq j\leq4$) denotes the index of a $4$-tuple coordinate vector for a cropping region. For convenience, each regressor $r_j^t$ is called as a primitive regressor. $T$ is the number of primitive regressors used for training. Note that the $4$-tuple coordinate vector consists of the coordinates of the top-left and bottom-right corners (i.e., $[\hat{x}_1,\hat{y}_1,\hat{x}_2,\hat{y}_2]$) of a rectangle region. Based on an initial cropping region $C^0$ and the obtained cropping-indexed CNN features $x^t$, each primitive regressor $r_j^t$ outputs a cropping region increment $\delta C_j^t$ (defined as $Y_j-C_j^{t-1}$), where $Y$ is the $4$-tuple coordinate vector for the cropping region annotated by a professional photographer. The estimates of the cascaded primitive regressors for image cropping are computed by accumulating the predicted values obtained by all the primitive regressors and the coordinates of an initial crop. The above process can be written as:
\begin{equation}
C_j=C_j^0+\sum_{t=1}^T{r_j^t(x^t)}.
\label{Eq::cascadedR} \end{equation}
Given $N$ training samples $(I_i,Y_i)_{i=1}^N$, the primitive regressors $(r_j^t(x^t))_{t=1}^T$ are sequentially learned until the test error stops decreasing. Each primitive regressor $r_j^t$ (an abbreviation of $r_j^t(x^t)$) is learned by minimizing the following objective function:
\begin{equation}
r_j^t=\mathop{\argmin}_{r_j}{\sum_{i=1}^N{\|Y_{ij}-(C_{ij}^{t-1}+r_j(x_i^t))}\|^2},
\label{Eq::optR} \end{equation}
where $Y_{ij}$ denotes the $j$th value in the $4$-tuple coordinate vector of the cropping region annotated by a professional photographer for the $i$th image.

The choice of primitive regressors is diverse and it mainly depends on what the application domain is. For example, in~\cite{Sun2013}, CNN is chosen as a primitive regressor to regress the coordinates of facial points. However, CNN is hard to be applied for the task of image cropping because of limited labelled cropping data.
In the task of pose regression~\cite{cascadedRe2010}, the random-ferns regressor, which shows better generalization ability than the tree regressor~\cite{cart93}, is proposed as the primitive regressor. However,  the experimental results in Section~\ref{sec::experiments} show that the random-ferns regressor is too weak for the task of cropping regression. Therefore, in this paper, we propose to learn the primitive regressor by using a group of random-ferns regressors to strengthen its regression ability. The gradient boosting algorithm~\cite{Friedman01} is employed to implement this. According to the gradient boosting algorithm \cite{Friedman01}, a non-parametric primitive regressor $P$ can be obtained by using the least-squares gradient boosting algorithm, which is given in Algorithm~\ref{alg:boost}.

\renewcommand{\algorithmicrequire}{\textbf{Initialization:}}
\renewcommand{\algorithmicensure}{\textbf{Iteration:}}
\begin{algorithm}[t]
  \footnotesize
  \caption{The Least-Squares Gradient Boosting Algorithm} \label{alg:boost}
\begin{algorithmic}[1] \STATE \textbf{Input}:
Data$(x_i,y_i)_{i=1}^N$;  The number of iterations $M$;  Initial value $g^0=0$
\FOR {$m=1, 2, \dots, M$}
  \STATE $e_i^m=y_i-g^{m-1}(x_i)$   ~~~~~~~~// compute the residual error
  \STATE $(\rho^m,\alpha^m)=\mathop{\argmin}_{\alpha,\rho}{\sum_{i=1}^N{[e_i^m-\rho h(x_i;\alpha)}]}$~//select a random-ferns regressor $h^m$ whose parameter is $\alpha^m$
  \STATE $g^m(x_i)=g^{m-1}(x_i)+h^m(x_i;\alpha^m)$ ~~~~~~//update the predicted value
  \ENDFOR
  \STATE \textbf{Output}: $P=(h^1,\cdots,h^M)$
\end{algorithmic}
\end{algorithm}

In Algorithm~\ref{alg:boost}, $g^m(x_i)$ denotes the predicted value for the $i$th sample in the $m$th iteration. $h(x_i;\alpha)$ denotes the random-ferns regressor and $\alpha$ denotes its parameter. $x_i$ denotes the features for
the $i$th image. $P=(h^1,\cdots,h^M)$ is the obtained primitive regressor, where $M$ is the number of the random-ferns regressors and $h^m$ is the random-ferns regressors with $\alpha^m$ as its parameter (i.e., $h^m(x_i;\alpha^m)$). $y_i$ is the given label value for the $i$th image and $\rho^m$ denotes the best gradient
descent step-size. In the implementation of the
random-ferns regressor, dozens of random-ferns regressors $h(x_i;\alpha)$ are generated uniformly
in the $m$th iteration and the random-ferns regressor $h^m(x_i;\alpha^m)$ with the best performance is chosen. Thus, the gradient descent step-size $\rho^m$ is an indicator vector with each element corresponding to one random-ferns regressor, where the value of an element is one if the
corresponding random-ferns regressor is chosen, and zero otherwise. The gradient boosting algorithm uses a weak regressor
$h(x_i;\alpha)$ to predict the residual error value $e_i^m$ between the predicted value $g^{m-1}(x_i)$ obtained from the ($m-1$)th iteration and the given label $y_i$.
The obtained regressor $P$ is also called as the master regressor consisting of
$M$ random-ferns regressors.  For more
details, please refer to~~\cite{Friedman01}.
\subsection{Cascaded Cropping Regression}
\renewcommand{\algorithmicrequire}{\textbf{Initialization:}}
\renewcommand{\algorithmicensure}{\textbf{Iteration:}}
\begin{algorithm}[t]
  \footnotesize
\caption{The Cascaded Cropping Regression Algorithm}
\label{alg::CCR}
\begin{algorithmic}[1]
    \STATE \textbf{Input}:
 Data$(I_i,Y_i)_{i=1}^N$;
 The number of stages $T$;
 Initial cropping region $(C_i^0)_{i=1}^N$
\FOR {$t=1, 2, \dots, T$}
  \STATE $x_i^t=f(C_i^{t-1},I_i)$  ~~~~~~// compute the cropping-indexed CNN features
  \FOR {$j=1, 2, \dots, 4$}
  \STATE $\tilde{y}_{ij}^t=Y_{ij}-C_{ij}^{t-1}$      ~~~~// compute the residual error
  \STATE Learning the primitive regressor $r_j^t$ with ($x_i^t$,$\tilde{y}_{ij}^t$) as the input by using Algorithm~\ref{alg:boost}
  \STATE $C_{ij}^t=C_{ij}^{t-1}+\lambda~r_j^t(x_i^t)$  ~~~~//update the cropping region
  \ENDFOR
 \ENDFOR
 \STATE \textbf{Output}: $R_j=(r_j^1,\cdots,r_j^T)_{j=1}^4$
\end{algorithmic}
\label{alg:nestedboost}
\end{algorithm}

\begin{figure}[]
\centering
   \includegraphics[width=0.88\linewidth]{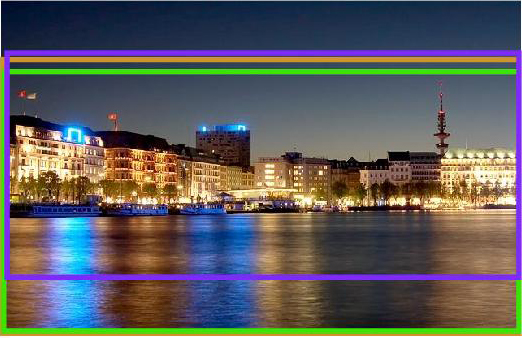}
   \caption{An example of the image cropping dataset provided by~\cite{Yan2013,Yan2015}. Three cropping regions are respectively annotated by three professional photographers for each image in the dataset. The three cropping regions are respectively shown in three different colors.}
\label{Fig:IOU3}
\end{figure}
\begin{figure}[]
\centering
   \includegraphics[width=0.86\linewidth]{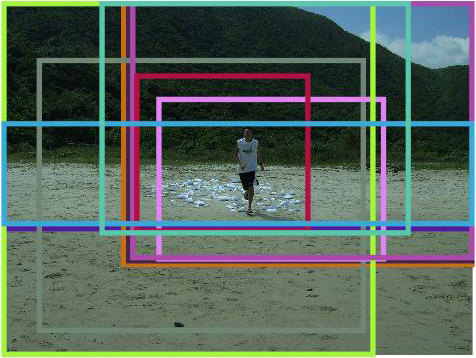}
   \caption{An example of the image cropping dataset provided by~\cite{Fang2014}. Ten cropping regions are respectively annotated by ten professional photographers for each image in the dataset. The ten cropping regions are respectively shown in ten different colors.}
\label{Fig:IOU10}
\end{figure}
This subsection describes the proposed CCR method in detail based on a set of random-ferns
regressors. A random-ferns regressor is derived from a random-ferns classifier, which is originally  proposed for classifying keypoints in \cite{FernClassify2010}. Assume that the number of weak classifiers in a random-ferns classifier is $M$, the random-ferns classifier randomly partitions
$F$-dimensional features into $M$ groups and the size of each group is $S$, where $S$ is a small
integer. Each weak classifier takes an $S$-dimensional feature vector in the
$S$-dimensional feature space (instead of the original $F$-dimensional space)
as the input. Thus, the total number of the parameters of each weak classifier is significantly reduced.
The random-ferns regressor is proposed in \cite{cascadedRe2010} for pose regression. Each random-ferns regressor also
takes an $S$-dimensional feature vector as an input and yields a real output
value. However, as mentioned in Section~\ref{sec::learingPR}, $S$-dimensional
features are weak for the task of cropping regression. Thus, we use a group of
random-ferns regressors to learn the primitive regressor by using the gradient boosting
algorithm. Each random-ferns regressor still randomly takes an $S$-dimensional
feature vector as an input, but the dimension of the features taken by the
obtained primitive regressor is more than $S$. Each primitive regressor indirectly
selects features from the $F$-dimensional cropping-indexed CNN features by aggregating
the features, which are selected by the group of random-ferns regressors. Since each random-ferns
regressor randomly selects an $S$-dimensional feature vector, some features may
be selected repeatedly. As a result, the total number of features selected by the primitive
regressor may vary. Based on the obtained primitive regressor, four CCR models are trained
to fit the values of the 4-tuple coordinate vectors corresponding to the cropping regions annotated by professional photographers, respectively. The whole procedure of the proposed CCR method is summarized in
Algorithm~\ref{alg::CCR}. In order to reduce the computational
complexity, the cropping-indexed CNN features are extracted in the outside loop and they are
kept unchanged in the inner loop.

\begin{figure*}[t]
\centering
   \centerline{\includegraphics[width=0.98\textwidth]{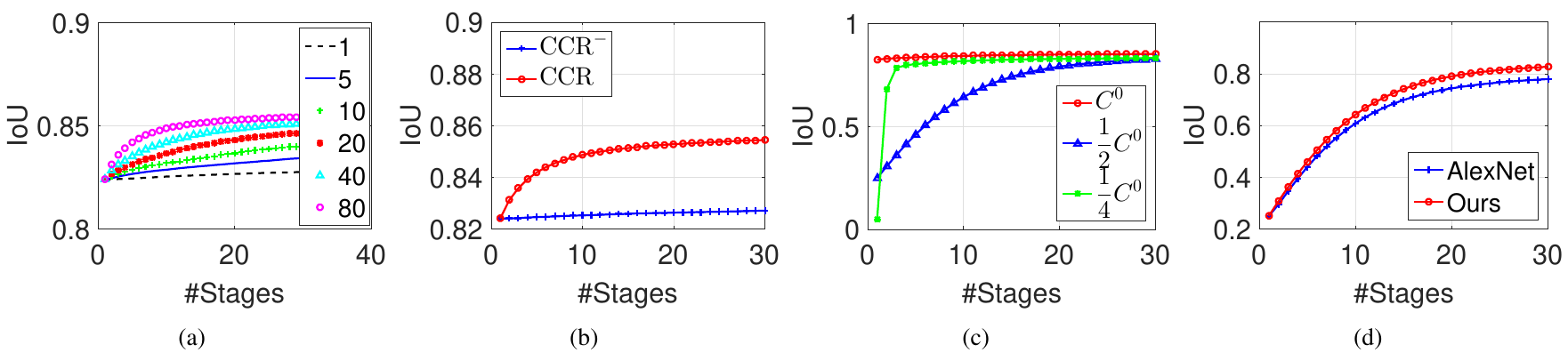}}

   \caption{\label{Fig::ParaCurves}The IoU curves obtained by the proposed method with different hyper-parameter settings. (a) The IoU curves with different numbers of random-ferns regressors. (b) The comparison of the convergence speed between the proposed CCR method and the CCR$^-$ method. (c) The IoU curves obtained by the proposed method with different initial cropping parameters. (d) The IoU curves obtained by the proposed method with different pre-trained CNN models.}
\end{figure*}
In Algorithm~\ref{alg::CCR}, in order to alleviate the problem of scale variations for different images, the coordinate values of the initial cropping region $C_i^0$ for the $i$th image are normalized with the maximum size of images.
$f(C_i^{t-1},I_i)$ denotes the procedure of extracting the cropping-indexed CNN features $x_i^t$ from the cropping region $C_i^{t-1}$ for the $i$th image $I_i$ at the $t$th stage. Then the residual error $\tilde{y}_{ij}^t$, which is the difference between the $j$th cropping value of the cropping region obtained at the ($t-1$)th stage and that of the cropping region annotated by a professional photographer, is computed. With ($x_i^t$,$\tilde{y}_{ij}^t$) as the input, the primitive regressor $r_j^t$ is learned by using Algorithm~\ref{alg:boost}. Then, the $j$th value of the cropping region $C_{ij}^t$ at the $t$th stage is updated with the predicted value obtained from $r_j^t$. The whole training process of CCR stops at the $T$th stage. After the training process, the obtained CCR model is used to predict a cropping region in an additive manner (see Eqn.~(\ref{Eq::cascadedR})).

The proof for the convergence of the proposed method is readily derived from \cite{Friedman01,BMR2002,cascadedRe2010}. Let the relative error obtained by the primitive regressor $P$ be  $\gamma=(\sum_i{d(r(x_i),y_i)}/\sum_i{d(y,y_i)})$, where $y$ denotes the single uniform prediction and $d$($\cdot$,$\cdot$) denotes a distance function.
The convergence of the proposed CCR method requires that $\gamma<1$, which means that the primitive regressor $P$ predicts a better result than the single uniform prediction.

\textbf{Regularization and Shrinkage.}
To prevent the proposed CCR method from overfitting, three regularization parameters are introduced in our case. As shown in \cite{Friedman01}, a regularization parameter for the gradient boosting models is the number of weak learners or regressors (i.e., the number of the random-ferns regressors $M$ in CCR). The best value of $M$ can be estimated by evaluating the performance of the CCR method on an independent validation set. In addition, controlling the value of $T$ is equivalent to control the training process to stop at the $T$th stage. Thus, $T$ is also a regularization parameter. The third regularization parameter is the shrinkage parameter, which often yields superior results to those obtained by restricting the number of regressors. In Algorithm~\ref{alg::CCR}, the introduction of the parameter $\lambda$ in line $7$ is a straightforward shrinkage strategy for scaling the update at different stages.
\section{Experiments}
\label{sec::experiments}

In this section, we evaluate the performance of the proposed image cropping method for visual aesthetic enhancement. The experiments include two parts. The first part evaluates the performance of the proposed method with different parameter settings. The second part evaluates the proposed method for image cropping and compares it with several state-of-the-art methods on cropping datasets. We use an image dataset, which contains 950 images collected by Yan~et~al.~\cite{Yan2013,Yan2015} from the CHUKPQ dataset, for evaluation. The image dataset contains seven classes of images, i.e., \emph{animal}, \emph{architecture}, \emph{human}, \emph{landscape}, \emph{night}, \emph{plant} and \emph{static}. A cropped region is respectively annotated for each image by three professional photographers, from which three cropping datasets are formed. Fig.~\ref{Fig:IOU3} shows an example of cropping regions annotated by the three professional photographers, where the regions outside of the bounding boxes are removed. Note that the images in the CHUKPQ dataset, which are the same as those in the cropping dataset, are removed in the process of training the aesthetic classifier.

We note that~\cite{Fang2014} contributes a new cropping dataset where each image has ten bounding boxes annotated by ten professional photographers. However, the ten annotated bounding boxes only have small overlaps (see Fig.~\ref{Fig:IOU10} for an example). It shows that there is little correlation among the ten professional photographers' knowledge or preference. Therefore, the proposed CCR method is not evaluated on that dataset.

Following the evaluation criteria used in \cite{Yan2013,Yan2015}, the Intersection over Union (IoU) metric and the Boundary Displacement Error (BDE) metric are adopted. The IoU metric is formulated as $IoU=\frac{W_p \bigcap W_m}{W_p \bigcup W_m}$
where $W_p$ and $W_m$ respectively denote a cropping window annotated by a professional photographer and a cropping window generated by an evaluated method on a test image. The BDE metric measures the Euclidean distance of the generated cropping boundaries $B_m$ from those annotated by the photographer $B_p$, which is written as $BDE=\frac{\|B_p-B_m\|^2}{4}$.

\begin{figure*}[t]
\centering
   \centerline{\includegraphics[width=0.98\textwidth]{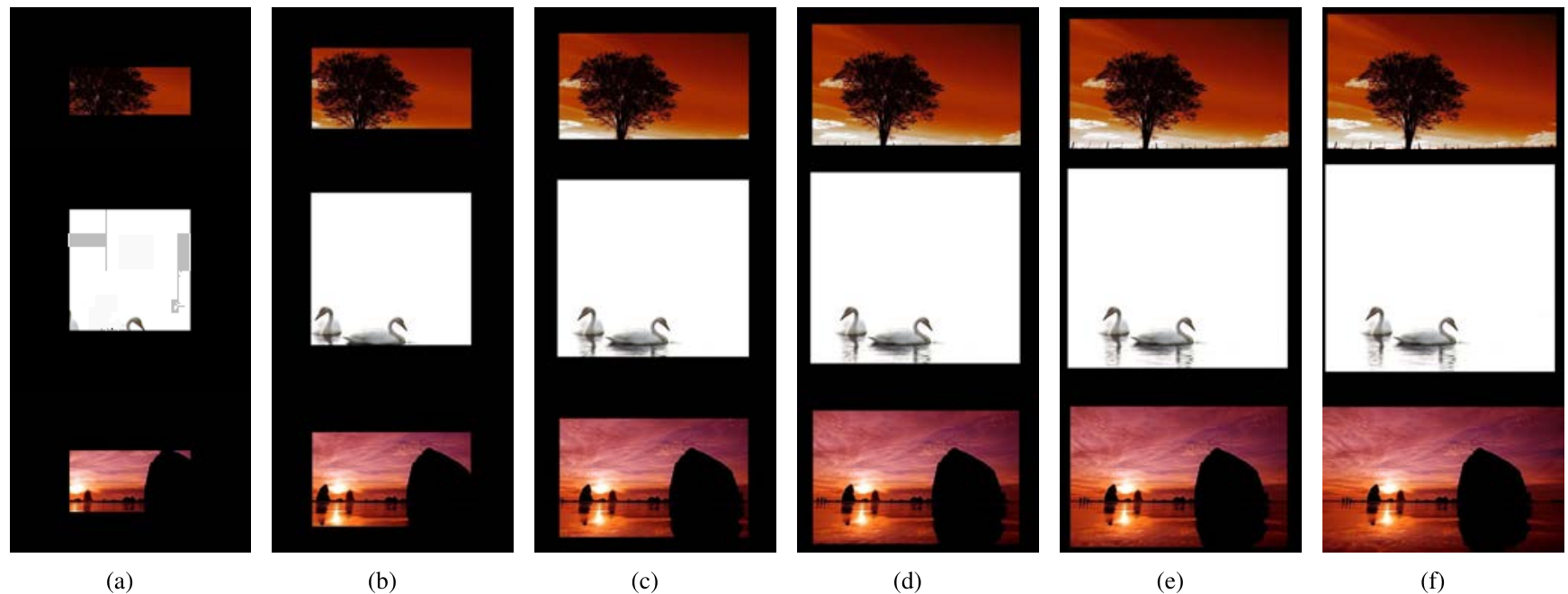}}
   \caption{\label{Fig::BoundaryCut}Examples showing that the proposed CCR method can alleviate the problem of boundary cutting. (a)-(f) show the results obtained by the proposed CCR method on three images at the 1st, 5th, 10th, 15th, 20th and 30th stage, respectively.}
\end{figure*}
\subsection{Influence of the Parameters}
After training CNN,  the obtained CNN model is applied to extract cropping-indexed CNN features for training a CCR model.
Several parameters are used for training the CCR model. In this subsection, we examine how these parameters affect the performance of the CCR method for image cropping.

\textbf{Influence of the number of random-ferns regressors.}
Fig.~\ref{Fig::ParaCurves}~(a) shows the IoU curves obtained by the proposed CCR method with different numbers of random-ferns regressors.
The convergence speed of the proposed CCR method is closely related to the number of random-ferns regressors. Usually, using more random-ferns regressors to learn a primitive regressor leads to faster convergence speed in the proposed CCR method. However, using too many random-ferns regressors to learn a primitive regressor may lead a CCR model to overfit a training dataset.

\textbf{Using different primitive regressors.}
To compare the convergence speed between the proposed CCR method and the cascaded method (i.e., the CPR method in~\cite{cascadedRe2010}), which uses a single random-ferns regressor as the primitive regressor, we apply the CPR method to the task of image cropping and compare its performance with that obtained by the proposed CCR method. In the implementation, the pose-indexed features in the CPR method are replaced by the cropping-indexed CNN features. We call this variant of the CPR method as CCR$^-$. The only difference between the proposed CCR method and the CCR$^-$ method is in that they use different primitive regressors. As shown in Fig.~\ref{Fig::ParaCurves} (b), the convergence speed of the proposed CCR method is significantly faster than that of the CCR$^-$ method. The proposed method converges in only about 30 stages, indicating that the proposed CCR method only extracts the cropping-indexed CNN features around 30 times to obtain the final cropping result. In contrast, the CCR$^-$ method only obtains a small improvement in  average IoU values in the first 30 stages. Thus, the proposed CCR method significantly improves the convergence speed when it uses multiple random-ferns regressors as a primitive regressor.

\textbf{Using different initial cropping values.}
The effect of the coordinates of an initial cropping region on the performance of the proposed CCR method for image cropping is also investigated. Each initial cropping region is simply scaled down to $50\%$ and $25\%$ of its original size without changing its centroid (i.e., $\frac{1}{2} C^0$ and $\frac{1}{4} C^0$ in Fig.~\ref{Fig::ParaCurves} (c)), respectively. Then the experiment is repeated again while the other parameters are fixed. The experimental results are shown in Fig.~\ref{Fig::ParaCurves} (c). As we can see that although the initial cropping parameters are different, the final cropping results are almost the same after several stages of regression. The proposed CCR method obtains the highest IoU values when the size of the initial cropping region is set to be the same as that of the original image. The reason is that the proposed CCR method mainly focuses on cutting out the regions from the current window, and more effective features can be selected by the proposed primitive regressor when the initial cropping region is larger than the cropping region to be estimated. In contrast, the proposed CCR method may infer image content outside of the current cropping box when the initial cropping region is smaller than the cropping region to be estimated, which makes the proposed CCR method less effective.  In addition, the coordinates of an initial cropping region significantly affect the convergence speed of the proposed CCR method. If the average IoU value at the initial stage is small, then the proposed CCR method has a fast convergence speed to increase the average IoU value. The differences in convergence speeds of the proposed method with different initial cropping parameters therefore become  smaller in the latter stages.

\textbf{Boundary cutting problem.} If the original initial cropping region is scaled down to 50\% or 25\% of its original size, then the cropping boundaries may pass through an object, which causes an unpleasant visual effect. However, the proposed method can alleviate this problem after several stages of regression. As shown in Fig.~\ref{Fig::BoundaryCut}, the boundaries of the tree are cut in the beginning stage due to the small size of the original initial cropping region. However, the proposed CCR method obtains a better cropping region after several stages of regression. Note that the above cropping region initialization strategy is very challenging, where the proposed CCR method tries to regress the bounding box for the image content outside the cropping region obtained from the previous stage. The proposed CCR method gradually crops images from inside to outside  to overcome the boundary cutting problem. However, the above cropping region initialization strategy is not optimal. As shown in Fig.~\ref{Fig::ParaCurves} (c), the proposed CCR method obtains the highest average IoU values when it crops images from outside to inside (i.e., the original image is used as the initial cropping region).

\textbf{Using different pre-trained CNN models to extract the cropping-indexed CNN features.} To show the effectiveness of the cropping-indexed CNN features extracted by the CNN model, which is trained on the aesthetic datasets, we use another pre-trained CNN model (AlexNet~\cite{alexNet}) to extract cropping-indexed CNN features for image cropping as a comparison. AlexNet is a pre-trained deep CNN model on the ImageNet dataset~\cite{ILSVRC15} for image classification and it includes 5 convolutional layers and 2 fully-connected layers. We empirically choose the output from the second fully-connected layer of AlexNet as the cropping-indexed CNN features. The second fully-connected layer of AlexNet is the best-performing layer for image cropping in our experiments. The experimental results are shown in Fig.~\ref{Fig::ParaCurves}~(d). We can see that the average IoU values obtained by the proposed CCR method using the  CNN model pre-trained on the aesthetic datasets are higher than those obtained by the proposed CCR method using the AlexNet CNN model. The reason is that the CNN model pre-trained on the aesthetic datasets can extract more effective aesthetic features, which are beneficial to visual aesthetic enhancement.
\begin{figure}
\centering
   \centerline{\includegraphics[width=0.48\textwidth]{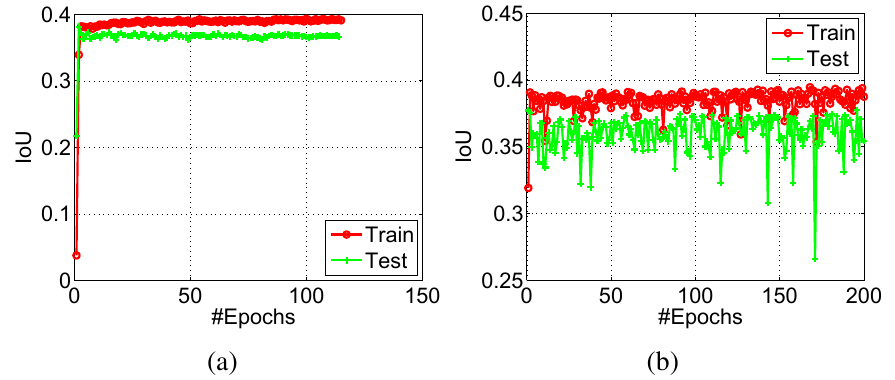}}
   \caption{\label{Fig::NativeCNN}The IoU curves obtained by a CNN regressor with different parameter initialization strategies on the training and test datasets. (a) The IoU curves obtained by the CNN regressor whose parameters are randomly initialized. (b) The IoU curves obtained by the CNN regressor, which copies the parameters from a  pre-trained CNN classifier, on the aesthetic datasets.}
\end{figure}
\begin{figure}
\centering
   \centerline{\includegraphics[width=0.48\textwidth]{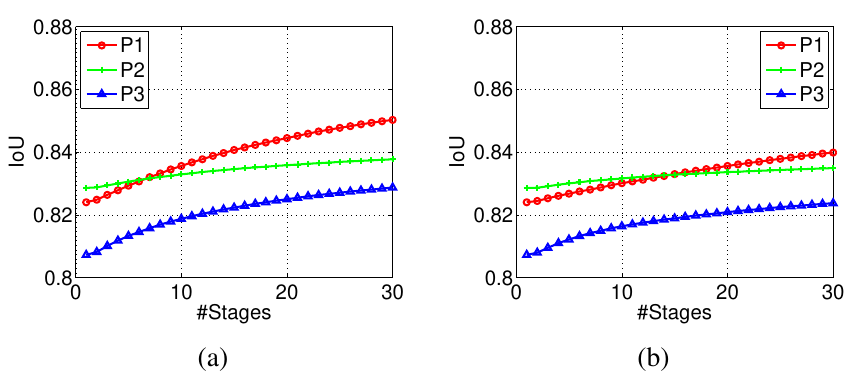}}
   \caption{\label{Fig:NCCRCurve}The IoU curves obtained by the proposed method on the three image cropping datasets with two different feature extraction strategies. (a) The IoU curves obtained by the proposed method with the cropping-indexed CNN features. (b) The IoU curves obtained by the proposed method with the mixed CNN features (i.e., the features extracted from the complete test images and the cropped regions obtained in each stage).}
\end{figure}

\begin{figure*}[t]
\centering
   \centerline{\includegraphics[width=0.98\textwidth]{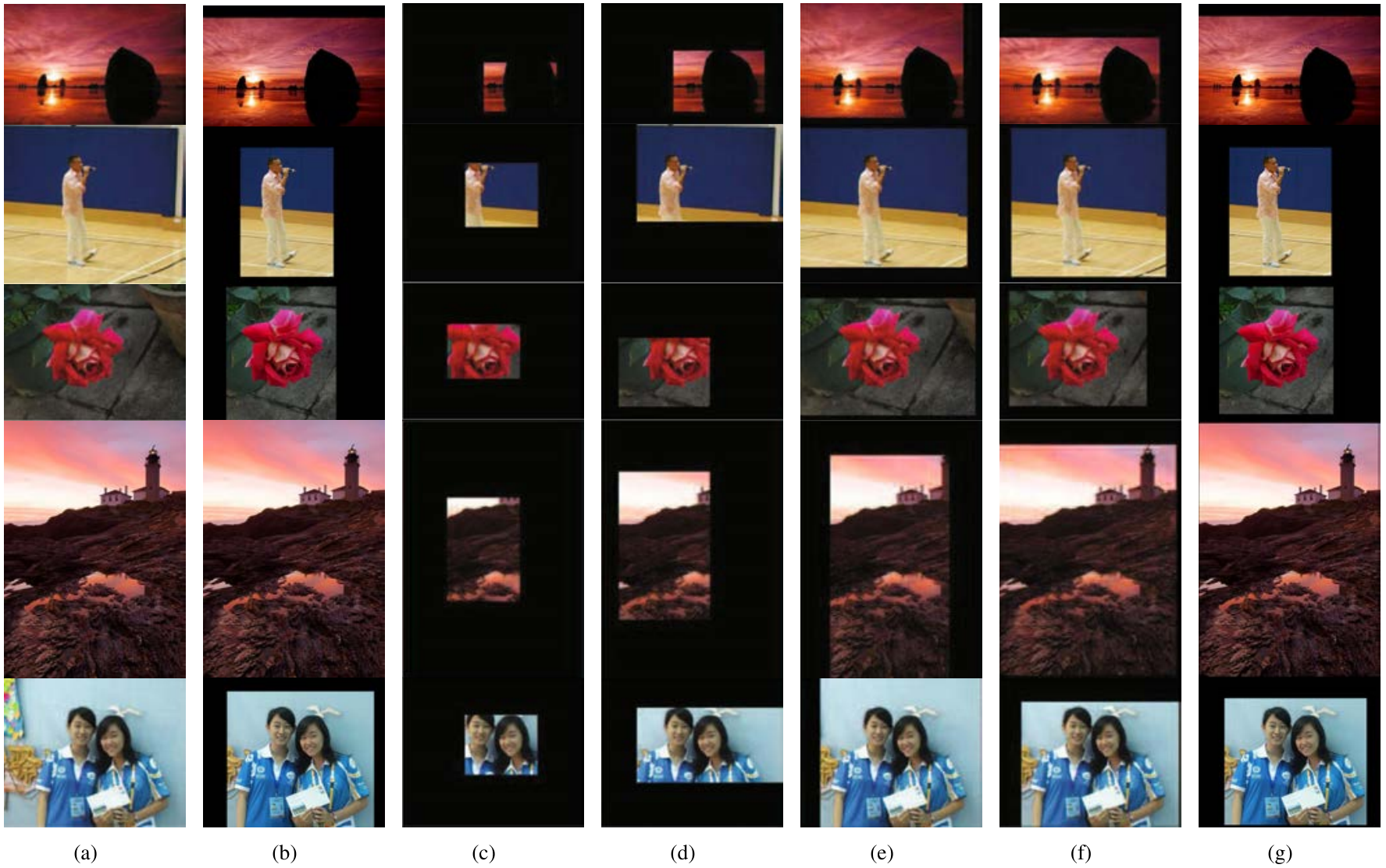}}
   \caption{\label{qualitativeComparing}Qualitative examples showing the cropped images obtained by the five competing methods. (a) The original Images. (b) The cropped images annotated by the first professional photographer. (c)-(g) The cropping results obtained by the attention-based~\cite{Stentiford_attentionbased}, the aesthetic-based~\cite{Nishiyama2009}, the change-based (2013)~\cite{Yan2013}, the change-based (2015)~\cite{Yan2015}, and the proposed methods.}
\end{figure*}
\textbf{Using a native CNN regressor with different parameter initialization strategies.} To show the effectiveness of the proposed CCR method, we also report the experimental results obtained by using a CNN regressor as a comparison. Two experiments are performed by the native CNN regressor with two different parameter initialization strategies to directly regress the  4-tuple coordinate vector. The CNN regressor has the same structure as that given in Fig.~\ref{Fig:CNNStructure} except that the number of the output nodes in the last layer is 4 instead of 2 in Fig.~\ref{Fig:CNNStructure}. In the first experiment, the parameters of the CNN regressor are randomly initialized, while they are initialized by using the parameters of a CNN  classifier pre-trained on the aesthetic datasets in the second experiment.  The experimental results are shown in Fig.~\ref{Fig::NativeCNN}. As can be seen, the results obtained by the CNN regressor in the two experiments are worse than those obtained by the proposed CCR method (see Fig.~\ref{Fig::ParaCurves}). The main reason is that only a limited number of training samples are used to train the CNN regressor, which affects the performance of the CNN regressor.
\subsection{Comparison with state-of-the-art methods}

In this subsection, we evaluate the performance of the proposed method for image cropping and compare it with several state-of-the-art methods on the cropping datasets. The comparison is performed using four state-of-the-art image cropping methods, which are the representative methods from three categories of image cropping methods (as discussed in Section~\ref{sec::relatedW}). The first two methods are the attention-based method~\cite{Stentiford_attentionbased} and the aesthetic-based method~\cite{Nishiyama2009}, respectively. Both methods are extended by Yan~et~al.~\cite{Yan2015}. For fairness, the extension version~\cite{Yan2015} of the two methods are used to achieve better results. The third method is the change-based method~\cite{Yan2013}, which learns the experience from professional photographers by using hand-crafted features. The improved version of the change-based method~\cite{Yan2015} is also used in this paper. The method in~\cite{Fang2014} is not evaluated for comparison since its source code is not available to the public.

\begin{table}
\begin{center}
\caption{\label{tab:ava}The test accuracy obtained by the pre-trained CNN classifier on the original cropping dataset and the three cropping datasets cropped by the proposed CCR method.}
\begin{tabular}{lc}
\hline
Datasets & Accuracy \\
\hline
original & 0.2512 \\
Photographer1 & 0.4370 \\
Photographer2 & 0.4141 \\
Photographer3 & 0.3815\\
\hline
\end{tabular}
\end{center}
\end{table}

\begin{table*}
\small
\begin{center}
\caption {\label{tblCom}The IoU (and BDE) results obtained by the four competing state-of-the-arts methods and the proposed method for image cropping. The results obtained by the other four competing methods are from~\cite{Yan2015}.}
\begin{tabular}{lccc}
\hline\noalign{\smallskip}
Methods & Photographer1 & Photographer2 & Photographer3\\
\noalign{\smallskip}
\hline
\noalign{\smallskip}
Attention-based~\cite{Stentiford_attentionbased}  & 0.203 (0.254) & 0.178 (0.200) &0.199 (0.259) \\
Aesthetic-based~\cite{Nishiyama2009}           & 0.396 (0.178)   & 0.394 (0.178) &0.386 (0.183) \\
Change-based (2013)~\cite{Yan2013}              & 0.749 (0.067)   & 0.729 (0.072) &0.732 (0.072)  \\
Change-based (2015)~\cite{Yan2015}              & 0.797 (0.053)   & 0.786 (0.057) &0.772 (0.059) \\
The proposed method                             & \textbf{0.850} (\textbf{0.032}) & \textbf{0.837} (\textbf{0.033}) &\textbf{0.828} (\textbf{0.035}) \\
\hline
\end{tabular}
\end{center}
\end{table*}

For the hyperparameter setting used for training the CCR model on the cropping
dataset~\cite{Yan2015} annotated by the first professional photographer, we set $S=4$, $M=20$ and $\lambda=0.800$. The images annotated by the other two professional photographers are used for test. This evaluation strategy is similar to that used in~\cite{Yan2013,Yan2015}. As shown in Fig.~\ref{Fig:NCCRCurve}(a), the IoU values obtained by the proposed method at the $30$th stage on the three cropping datasets are $0.850$, $0.837$ and $0.828$, respectively. However, we note that the proposed CCR can also achieve promising results even at the 10th stage, where the IoU values obtained by the proposed CCR method on the three datasets are respectively $0.839$, $0.834$ and $0.823$.

To show the effectiveness of the proposed cropping-indexed CNN features, we evaluate the performance of the proposed method with the mixed CNN features that are simultaneously extracted from both a complete test image and the cropped region for comparison. In each stage, the pre-trained CNN model is respectively applied on the complete test image and the cropped region obtained from the previous stage, to extract the features. The features obtained  from both the complete test image and the cropped region are concatenated as a vector, which is used as the mixed CNN features. The experimental results are shown in Fig.~\ref{Fig:NCCRCurve}. As can be seen, the IoU values obtained by the proposed method with the mixed features  at the 30th stage on the three cropping datasets are respectively 0.838, 0.834 and 0.823, which are lower than those obtained by the proposed method with only the cropping-indexed CNN features. The reason is that  the cropping-indexed CNN features can provide aesthetic discriminative information. Moreover, the proposed primitive regressor can select the features, which are beneficial to image cropping for visual aesthetic enhancement. In contrast, the mixed CNN features include many redundant features, which will lead to high computational time and the overfitting problem. More specifically, effective features are hard to be selected from the mixed CNN features by the proposed primitive regressor because that the length of the mixed CNN feature vector is higher than that of the cropping-indexed CNN feature vector. Thus, the mixed CNN features are less effective than the cropping-indexed CNN features for cropping region regression. Additionally, the pre-trained CNN classifier is used to classify the images from the original cropping dataset and the cropped images obtained by the proposed CCR method, respectively. The classify accuracies are shown in Table~1. As can be seen, the obtained classification accuracies on the cropped images are higher than those on the original images about 13\%-18\%, which shows the effectiveness of the proposed image cropping method.

The results obtained by the proposed method are compared with those obtained by the other four competing methods. Table~\ref{tblCom} lists the comparison results. Some examples of the cropping regions annotated by the first professional photographer and the cropping regions obtained by the five competing methods are also shown in Fig.~\ref{qualitativeComparing}. As we can see from Table~\ref{tblCom} and Fig.~\ref{qualitativeComparing}, the proposed CCR method clearly outperforms the other four competing methods, and the cropping regions obtained by the proposed CCR method are closer to those annotated by the professional photographers than the other competing methods. The change-based method~\cite{Yan2013} and its improved version~\cite{Yan2015} are similar to the proposed CCR method because they also learn the experience from professional photographers. However, due to the fact that the hand-crafted features used in the change-based method and its improved version are often limited, both methods obtain worse performance than the proposed method. Among the five competing methods, the attention-based method~\cite{Stentiford_attentionbased} obtains the worst IoU and BDE results on all the three cropping datasets since it only focuses on the salient objects in an image while it ignores the aesthetic of a cropped image. The aesthetic-based method~\cite{Nishiyama2009} achieves better results than the attention-based method~\cite{Stentiford_attentionbased}, but it performs worse than the change-based method. This is because that the aesthetic-based method focuses more on high-quality local cropping regions, while it may ignore the object regions in an image. The proposed method and the change-based methods can overcome this drawback by learning the knowledge of the professional photographers. Thus, they can obtain better results than the aesthetic-based method.

\begin{figure}
\begin{center}
   \centerline{\includegraphics[width=0.48\textwidth]{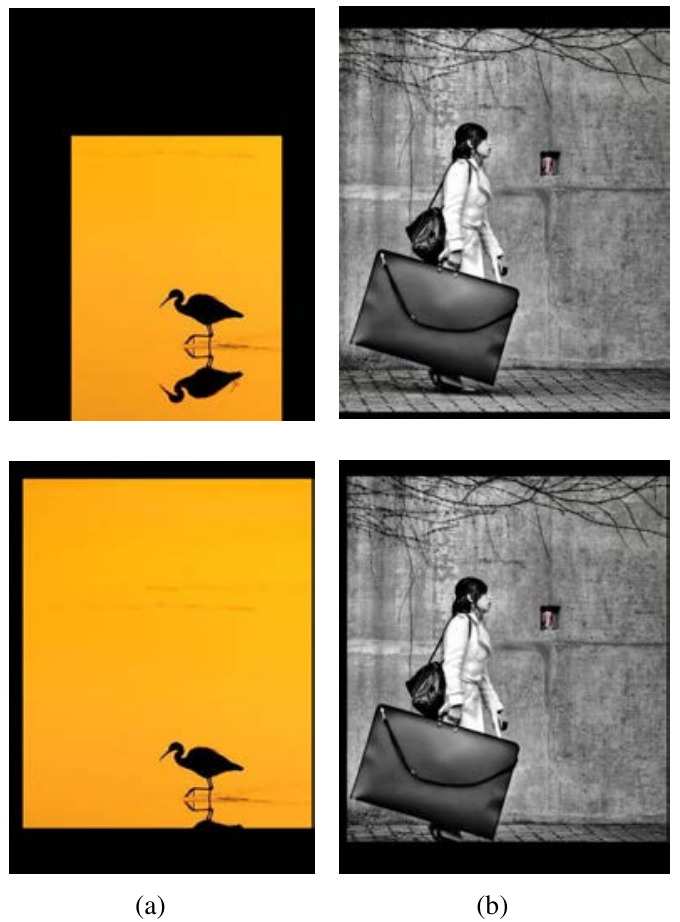}}
\end{center}
   \caption{\label{failureExamples} Failure examples obtained by the proposed method. From top to bottom, each row shows the cropping results annotated by the first professional photographer and the cropping results obtained by the proposed method, respectively.}
\end{figure}

To analyze the failure case of the proposed method, we also give some failure examples (see Fig.~\ref{failureExamples}) obtained by the proposed method. Two kinds of images may cause the failure of the proposed method. The first kind of images are those in which the position of an object is close to the image boundaries. The proposed method may cut off a small part of the object in an image if the regression results are not accurate enough. As shown in the first failure example in Fig.~\ref{failureExamples}(a), a part of the reflection of the bird is removed by the proposed method. The second kind of images usually have no much texture in their backgrounds. As shown in the first failure example in the figure, the background of the example is a solid color. The proposed method may not work quite well for the second kind of images. The total numbers of failure cases obtained by the proposed CCR method on the three cropping datasets are listed in Table~\ref{tab:failure}. As can be seen, the proposed CCR method only fails on a small percentage of images (4.3\% - 6.3\% of the whole test images).

\begin{table}
\small
\begin{center}
\caption {\label{tab:failure}The numbers of failure cases (denoted by \#Failure) obtained by the proposed CCR method in the cropping datasets.}
\begin{tabular}{lccc}
\hline\noalign{\smallskip}
Dataset & Photographer1 & Photographer2 & Photographer3\\
\noalign{\smallskip}
\hline
\noalign{\smallskip}
\#Failure  & 41 & 60 &56 \\
\hline
\end{tabular}
\end{center}
\end{table}

In terms of running time, the total execution time of the proposed method (implemented in Matlab on a 4 GHz 32GB RAM PC) on each image is about 1.7 seconds. Most running time is used to extract the cropping-indexed CNN features. In contrast, the running time of the improved change-based method (implemented in C++ on a 2.33 GHz 4GB RAM PC) is about 11 seconds~\cite{Yan2015}.
\section{Conclusion}
In this paper, we propose a novel CCR method to perform automatic image cropping for visual aesthetic enhancement based on the cropping-indexed CNN features. A two-step learning strategy is proposed for a limited number of labelled cropping data. In addition, the proposed CCR method learns a primitive regressor by using the gradient boosting strategy on a group of random-ferns regressors at each stage. The convergence speed of the proposed CCR method is much faster than that of the cascaded regression method directly using random-ferns regressors. On the other hand, the performance of the proposed CCR method for image cropping also benefits from the cropping-indexed CNN features, which are extracted by the CNN model trained on large-scale visual aesthetic datasets. Experimental results show that the proposed method is quite effective and efficient for image cropping.

\section*{Acknowledgments} This work was supported by the National Natural Science Foundation of China under Grants U1605252, 61472334 and 61571379, and by the Natural Science Foundation of Fujian Province of China under Grant 2017J01127.


%

%
%
%
%
%

\ifCLASSOPTIONcaptionsoff
  \newpage
\fi



%
%
%

\bibliographystyle{IEEEtran}
\bibliography{ImgCropping}

%

\begin{IEEEbiography}[{\includegraphics[width=1in,height=1.25in,clip,keepaspectratio]{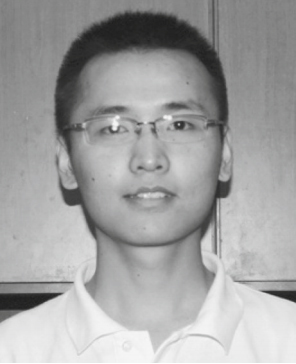}}]{Guanjun Guo}
is currently pursuing the Ph.D. degree with the Fujian Key Laboratory of Sensing and Computing for Smart City, and the School of Information Science and Engineering, Xiamen University, Xiamen, China.
His current research interests include computer vision, machine learning.
\end{IEEEbiography}
\begin{IEEEbiography}[{\includegraphics[width=1in,height=1.25in,clip,keepaspectratio]{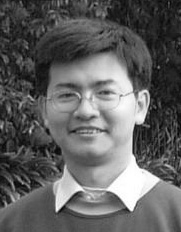}}]{Hanzi Wang} (SM'10)
is currently a Distinguished Professor of §Minjiang Scholars§ in Fujian province and a Founding Director of the Center for Pattern Analysis and Machine Intelligence (CPAMI) at XMU. He was an Adjunct Professor (2010-2012) and a Senior Research Fellow (2008-2010) at the University of Adelaide, Australia; an Assistant Research Scientist (2007-2008) and a Postdoctoral Fellow (2006-2007) at the Johns Hopkins University; and a Research Fellow at Monash University, Australia (2004-2006). He received his Ph.D degree in Computer Vision from Monash University where he was awarded the Douglas Lampard Electrical Engineering Research Prize and Medal for the best PhD thesis in the department. His research interests are concentrated on computer vision and pattern recognition including visual tracking, robust statistics, object detection, video segmentation, model fitting, etc. He has published more than 100 papers in major international journals and conferences including the IEEE T-PAMI, IJCV, ICCV, CVPR, ECCV, NIPS, MICCAI, etc.
He is a senior member of the IEEE. He was an Associate Editor for IEEE Transactions on Circuits and Systems for Video Technology (from 2010 to 2015) and a Guest Editor of Pattern Recognition Letters (September 2009). He was the General Chair for ICIMCS2014, Program Chair for CVRS2012, Publicity Chair for IEEE NAS2012, and Area Chair for ACCV2016, DICTA2010. He also serves on the program committee (PC) of ICCV, ECCV, CVPR, ACCV, PAKDD, ICIG, ADMA, CISP, etc, and he serves on the reviewer panel for more than 40 journals and conferences.
\end{IEEEbiography}
\begin{IEEEbiography}[{\includegraphics[width=1in,height=1.25in,clip,keepaspectratio]{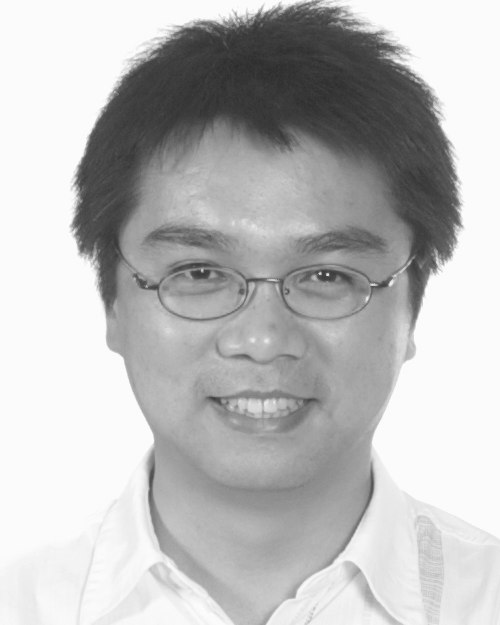}}]{Chunhua Shen}
is a Professor at School of Computer Science, University of Adelaide. He is a Project Leader and Chief Investigator at the Australian Research Council Centre of Excellence for Robotic Vision (ACRV), for which he leads the project on machine learning for robotic vision. Before he moved to Adelaide as a Senior Lecturer, he was with the computer vision program at NICTA (National ICT Australia), Canberra Research Laboratory for about six years. His research interests are in the intersection of computer vision and statistical machine learning. Recent work has been on real-time object detection, large-scale image retrieval and classification, and scalable nonlinear optimization.
He studied at Nanjing University, at Australian National University, and received his PhD degree from the University of Adelaide. From 2012 to 2016, he holds an Australian Research Council Future Fellowship. He is serving as Associate Editor of IEEE Transactions on Neural Networks and Learning Systems.
\end{IEEEbiography}
\begin{IEEEbiography}[{\includegraphics[width=1in,height=1.25in,clip,keepaspectratio]{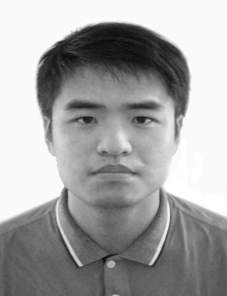}}]{Yan Yan}
is currently an associate professor in the School of Information Science and Technology at Xiamen University, China. He received the Ph.D. degree in Information and Communication Engineering from Tsinghua University, China, in 2009. He worked at Nokia Japan R\&D center as a research engineer (2009-2010) and Panasonic Singapore Lab as a project leader (2011). He has published around 40 papers in the international journals and conferences including the IEEE T-IP, T-Cyber, T-ITS, PR, KBS, Neurocomputing, ICCV, ECCV, ACM MM, ICPR, ICIP, etc. His research interests include computer vision and pattern recognition.
\end{IEEEbiography}
\begin{IEEEbiography}[{\includegraphics[width=1in,height=1.25in,clip,keepaspectratio]{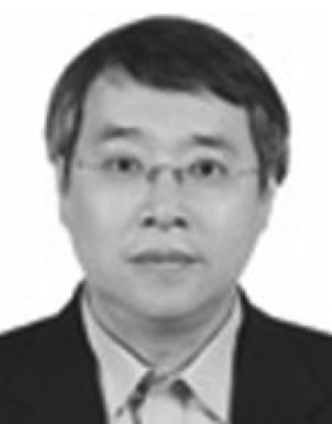}}]{Hong-Yuan Mark Liao} (F'13) received the PhD degree in electrical engineering from Northwestern
University, Evanston, IL, in 1990. In 1991, he joined the Institute of Information Science, Academia Sinica, Taipei, Taiwan, where he is currently a distinguished research fellow. He was in the fields of multimedia signal processing, image processing, computer vision, pattern recognition, video forensics, and multimedia protection for more than 25 years. He received the Young Investigators Award from Academia Sinica in 1998, the Distinguished Research Award from the National Science Council of Taiwan, in 2003, 2010, and 2013, respectively, the National Invention Award of Taiwan in 2004, the Distinguished Scholar Research Project Award from the National Science Council of Taiwan in 2008, and the Academia Sinica Investigator Award in 2010. His professional activities include the cochair of the 2004 International Conference on Multimedia and Exposition (ICME), the Technical cochair of the 2007 ICME, the General cochair of the 17th International Conference on Multimedia Modeling, the President of the Image Processing and Pattern Recognition Society of Taiwan (2006每2008), an Editorial Board member of the IEEE Signal Processing Magazine (2010每2013), and an associate editor of the IEEE Transactions On Image Processing (2009每2013), the IEEE Transactions On Information Forensics And Security (2009每2012), and the IEEE Transactions On Multimedia (1998每2001). He also serves as the IEEE Signal Processing Society Region 10 Director (Asia-Pacific Region). He is a fellow of the IEEE.
\end{IEEEbiography}




\end{document}